\documentclass[11pt,letterpaper]{article}
\usepackage{emnlp2017}
\usepackage{times}
\usepackage{latexsym}
\usepackage{enumitem}
\usepackage{url}
\usepackage{hyperref}
\usepackage{pgfplots}
\usepackage{latexsym}
\usepackage{tikz}
\usepackage{booktabs}
\usepackage{graphicx}
\usepackage{subfigure}
\usepackage{multirow}
\usepackage{amsmath}
\usepackage{amsfonts}
\usepackage{graphics}
\usepackage[british]{babel}
\usepackage{mathrsfs}
\usepackage{makecell}

\definecolor{yellow2}{RGB}{166,97,26}
\definecolor{yellow1}{RGB}{223,194,125}
\definecolor{orange1}{RGB}{239,138,98}

\definecolor{gray_bg}{RGB}{230,230,230}
\definecolor{gray3}{RGB}{240,240,240}
\definecolor{gray2}{RGB}{153,153,153}
\definecolor{gray1}{RGB}{77,77,77}
\definecolor{pink3}{RGB}{178,50,50}
\definecolor{pink2}{RGB}{239,138,98}
\definecolor{pink1}{RGB}{102,194,164}
\definecolor{white1}{RGB}{245,245,245}
\definecolor{green4}{RGB}{0,109,44}
\definecolor{green3}{RGB}{220,180,200}
\definecolor{green2}{RGB}{103,169,207}
\definecolor{green1}{RGB}{33,102,172}
\definecolor{dark_cyan}{rgb}{0.0, 0.55, 0.55}
\definecolor{purple1}{rgb}{0.69,0.61,0.85}
\definecolor{purple2}{rgb}{0.47,0.32,0.66}

\emnlpfinalcopy

\def\emnlppaperid{***}

\title{Spherical Paragraph Model}

\author{Ruqing Zhang, Jiafeng Guo, Yanyan Lan, Jun Xu\& Xueqi Cheng\\
CAS Key Lab of Network Data Science and Technology\\
Institute of Computing Technology, Chinese Academy of Sciences\\
Beijing, China \\
\texttt{zhangruqing@software.ict.ac.cn, \{guojiafeng,lanyanyan,junxu,cxq\}@ict.ac.cn} \\
}

\date{}

\begin{document}

\maketitle
\begin{abstract}
Representing texts as fixed-length vectors is central to many language processing tasks. Most traditional methods build text representations based on the simple Bag-of-Words (BoW) representation, which loses the rich semantic relations between words. Recent advances in natural language processing have shown that semantically meaningful representations of words can be efficiently acquired by distributed models, making it possible to build text representations based on a better foundation called the Bag-of-Word-Embedding (BoWE) representation. However, existing text representation methods using BoWE often lack sound probabilistic foundations or cannot well capture the semantic relatedness encoded in word vectors. To address these problems, we introduce the Spherical Paragraph Model (SPM), a probabilistic generative model based on BoWE, for text representation. SPM has good probabilistic interpretability and can fully leverage the rich semantics of words, the word co-occurrence information as well as the corpus-wide information to help the representation learning of texts. Experimental results on topical classification and sentiment analysis demonstrate that SPM can achieve new state-of-the-art performances on several benchmark datasets.
\end{abstract}

\section{Introduction}

A central question to many language understanding problems is how to capture the essential meaning of a text in a machine-understandable format (\textit{e.g.}, fixed-length vector representation). Most traditional methods either directly use the Bag-of-Words (BoW) representation \cite{harris1954distributional}, or built upon BoW using matrix factorization \cite{deerwester1990indexing,lee1999learning} or probabilistic topical models \cite{hofmann1999probabilistic,blei2003latent}. However, by using BoW as the foundation, rich semantic relatedness between words is lost. The text representation thus is obtained/learned purely based on the word-by-text co-occurrence information. However, humans understand a piece of text not solely based on its content (\textit{i.e.}, the word occurrences), but also her background knowledge (\textit{e.g.}, semantics of the words). Recent advances in the Natural Language Processing (NLP) community have shown that semantics of the words or more formally the distances between the words can be effectively revealed by distributed word representations \cite{mikolov2013efficient}, also referred to as ``word embeddings'' or ``word vectors''. Therefore, a natural idea is that one can build text representations based on a better foundation, namely the Bag-of-Word-Embeddings (BoWE) representation,  by replacing distinct words with word vectors learned a priori with rich semantic relatedness encoded.

There have been some recent attempts to use BoWE for text representations.  Perhaps the simplest way is to represent the text as a weighted average of all its word vectors \cite{vulic2013cross}. Besides, \citeauthor{clinchant2013aggregating} \shortcite{clinchant2013aggregating} aggregated the word vectors into a text-level representation under the Fisher Kernel framework. Another well-known approach is the Paragraph Vector (PV) \cite{le2014distributed}, which jointly learns the word and text representations as a direct optimization problem. There are several clear drawbacks with existing methods: (1) Existing methods often lack sound probabilistic foundations, making them heuristic or weak in interpretability; (2) All the methods assume the independency between texts, limiting their ability to leverage the corpus-wide information to help the representation learning of each piece of text. This limitation is analogous to that of Probabilistic Latent Semantic Indexing (PLSI) \cite{hofmann1999probabilistic} in topic modeling, which has been addressed by Latent Dirichlet Allocation (LDA) \cite{blei2003latent}; (3) Simple weighted sum or aggregation using fisher kernel cannot well capture the semantic relatedness encoded in word vectors, which is typically revealed by the distance (or similarity) between word vectors.

To address these problems, we introduce a novel Spherical Paragraph Model (SPM), which learns text representations through modeling the generation of the corpus based on BoWE representations. Specifically, each piece of text is first represented as a bag of $\ell_2$-normalized word vectors. Note that by normalization, the cosine similarity between word vectors are equal to the dot product between them, and all the word vectors lie on a unit hypersphere. We then assume the following generation process of the whole corpus. A text vector is first sampled from a corpus-wide prior distribution, and a word vector is then sampled from a text-level distribution given the text vector. The von Mises-Fisher (vMF) distribution \cite{banerjee2005clustering} is employed for both corpus-wide and text-level distributions, which arises naturally for data distributed on the unit hypersphere and model the directional relation (\textit{i.e.}, dot product) between vectors. The text representations can then be inferred by maximizing the likelihood of the generation of the whole corpus. We develop a variational EM algorithm to learn the SPM efficiently.

Compared with previous methods, SPM enjoys the following merits: (1) By modeling the generation process of the whole corpus based on BoWE, SPM can fully leverage the rich semantics of words, the word-by-text co-occurrences information as well as the corpus-wide information to help the representation learning of texts; (2) By employing the vMF distribution, SPM can well capture the semantic relatedness encoded in words vectors (\textit{i.e.}, cosine similarity between word vectors); (3) SPM has good probabilistic interpretability as traditional topic models (\textit{e.g.}, LDA), while allows unlimited hidden topics (\textit{i.e.}, word clusters) as neural embedding models (\textit{e.g.}, PV) by eliminating the topic layer.

We evaluated the effectiveness of our SPM by comparing with existing text presentation methods based on several benchmark datasets. The empirical results demonstrate that our model can achieve new state-of-the-art performances on several topical classification and sentiment analysis tasks.

\section{Related Work}
\label{background}
In this section, we briefly review the existing text representation methods, and text models using the vMF distribution.

\subsection{Existing models for Texts}

The most common fixed-length representation is Bag-of-Words (BoW) \cite{harris1954distributional}. For example, in the popular TF-IDF scheme \cite{salton1986introduction}, each text is represented by \textit{tfidf} values of a set of selected feature-words. However, the BoW representation often suffers from data sparsity and high dimension. Meanwhile, by viewing each word as a distinct feature dimension, the BoW representation has very little sense about the semantics of the words.

To address this shortcoming, several dimensionality reduction methods have been proposed based on BoW, including matrix factorization methods such as LSI \cite{deerwester1990indexing} and NMF \cite{lee1999learning}, and probabilistic topical models such as PLSI \cite{hofmann1999probabilistic} and LDA \cite{blei2003latent}. The key idea of LSI is to map texts to a vector space of reduced dimensionality (\textit{i.e.}, the latent semantic space), based on a Singular Value Decomposition (SVD) over the term-document co-occurrence matrix. NMF is distinguished from the other methods by its non-negativity constraints, which leads to a parts-based representation because they allow only additive, not subtractive combinations. In PLSI, each word is generated from a single topic, and different words in a document may be generated from different topics. LDA is proposed by introducing a complete generative process over the documents, and demonstrated as a state-of-the-art document representation method.
However, as built upon the BoW representation, all these methods do not leverage the rich semantics of the words, and learn the text representations purely based on the word-by-text co-occurrence information.

Recent developments in distributed word representations have succeeded in capturing semantic regularities in language. Specifically, neural embedding models, \textit{e.g.}, Word2Vec model \cite{mikolov2013efficient} and Glove model \cite{pennington2014glove}, learn word vectors (also called word embeddings) efficiently from very large text corpus. The learned word vectors can reveal the semantic relatedness between words and perform word analogy tasks successfully.

With rich semantics encoded in word vectors, a natural question is how to obtain the text representation based on word vectors. A simple approach is to use a weighted average \cite{clinchant2013aggregating} or sum of all the word vectors. Besides, Fisher Vector (FV) \cite{clinchant2013aggregating} transforms the variable-cardinality word vectors into a fixed-length text representation based on the Fisher kernel framework \cite{jaakkola1999exploiting}. However, these methods often lack sound probabilistic foundations. Meanwhile, simple weighted sum or aggregation using fisher kernel cannot well capture the semantic relatedness encoded in word vectors, which is typically revealed by the distance (or similarity) between word vectors.
Later, Paragraph Vector (PV) which has two different model architectures (\textit{i.e.}, PV-DM and PV-DBOW) \cite{le2014distributed} is introduced to jointly learn the word and text representations. 
Although these models seem to work well in practice, there is a strong independence assumption between texts in these methods, limiting their ability to leverage the corpus-wide information to help the representation learning of each piece of text.

Besides these unsupervised representation learning methods, there have been many supervised deep models which directly learn text representations for the prediction tasks. Recursive Neural Network \cite{socher2013recursive} has been proven to be efficient in terms of constructing sentence representations. Recurrent Neural Network \cite{sutskever2011generating} can be viewed as an extremely deep neural network with weight sharing across time. Convolution Neural Network \cite{kim2014convolutional} can fairly determine discriminative phrases in a text with a max-pooling layer. However, these deep models are usually task dependent and time-consuming in training due to the complex model structures.

\subsection{vMF in Text Models}

The von Mises-Fisher distribution is known in the literature on directional statistics \cite{fisher1953dispersion,jupp1989unified,mardia2009directional}, and suitable for data distributed on the unit hypersphere. Here we first review the vMF distribution.

A d-dimensional unit random vector $x$ (\textit{i.e.}, $x \in \mathbb{R}^K$ and $||x|| = 1$) is said to have $K$-variate von Mises-Fisher distribution if its probability density function is given by,
\begin{equation*}
   f(x|\mu,\kappa){=} c_K{(\kappa)}  e^{\kappa {\mu}^\mathrm{T} x}
\end{equation*}
where $||\mu||=1$, $\kappa \ge 0$ and $K \ge 2$. The normalizing constant $c_K{(\kappa)}$ is given by,
\begin{equation*}
   c_K{(\kappa)} {=} \frac{{\kappa}^{K/2-1}}{(2\pi)^{K/2} I_{K/2-1}{(\kappa)}}
\end{equation*}
where $I_r(\cdot)$ represents the modified Bessel function of the first kind and order $r$. The density $f(x|\mu,\kappa)$ is parameterized by the mean direction $\mu$, and the concentration parameter $\kappa$. The concentration parameter $\kappa$ characterizes how strongly the unit vectors drawn from the distribution are concentrated on the mean direction $\mu$.

The vMF distribution has properties analogous to those of the multi-variate Gaussian distribution for data in $\mathbb{R}^{K}$, parameterized by cosine similarity rather than Euclidean distance. Evidence suggests that this type of directional measure (\textit{i.e.}, cosine similarity) is often superior to Euclidean distance in high dimensions \cite{manning1999foundations,zhong2005generative}.

The vMF distribution has been applied in text representations based on BoW in literature. For example, \citeauthor{banerjee2005clustering} \shortcite{banerjee2005clustering} introduced the mixture of von Mises-Fisher distributions (movMF) that serves as a generative model for directional text data. The movMF model treats each normalized text vector (\textit{i.e.}, normalized \textit{tf} or \textit{tf-idf} vector) as drawn from one of the $M$ vMF distributions centered on one cluster mean, selected by a mixing distribution. The cluster assignment variable for instance $x_i$ is denoted by $z_i \in \{1,2,\dots, M\}$. The probabilistic generative process is given by,
\vskip -0.1in
\begin{displaymath}
\begin{array}{rcl}
z_i & \sim & \text{Categorical}(.|\pi)\\
x_i & \sim & \text{vMF}(.|\mu_{z_i},\kappa)
\end{array}
\end{displaymath}
where parameters $\Theta=\{\pi,\boldsymbol{\mu},\boldsymbol{\kappa}\}$ are treated as fixed unknown constants and $\boldsymbol{Z}={\{z_i\}}_{i=1}^{M}$ are treated as a latent variables.

Later, \citeauthor{reisinger2010spherical} \shortcite{reisinger2010spherical} introduced the Spherical Admixture Model (SAM), a Bayesian admixture model of normalized vectors on $\mathbb{S}^{K-1}$.
The generative model is given by,
\vskip -0.3in
\begin{displaymath}
\begin{array}{rcl}
\mu|\kappa_0 & \sim & $vMF$(m,\kappa_0)\\
\phi_t|\mu,\xi & \sim & $vMF$(\mu,\xi) \\
\theta_d|\alpha & \sim & $Dirichlet$(\alpha) \\
\bar{\phi}_d|\phi,\theta_d & = & $Avg$(\phi,\theta_d) \\
v_d|\bar{\phi}_d,\kappa & \sim & $vMF$(\bar{\phi}_d,\kappa)\\
\end{array}
\end{displaymath}
where $\mu$ is the corpus mean direction, $\xi$ controls the concentration of topics around $\mu$, the elements of $\theta_d$ are mixing proportions for text $d$, and $v_d$ is the observed vector for text $d$.

All these vMF-based methods treat the text as a single object (\textit{i.e.}, a normalized feature vector), and successfully integrate a directional measure of similarity into a probabilistic setting for text modeling. However, the foundations of these methods are still BoW, which means that they cannot leverage the rich semantic relatedness between the words for text representation. Unlike these methods, we use vMF to capture the semantic relatedness encoded in word vectors revealed by cosine similarity, and build text representations based on a better BoWE foundation.

\section{Spherical Paragraph Model}
\label{SPM}

\begin{figure}[t]
	\centering
		\includegraphics[scale=0.32]{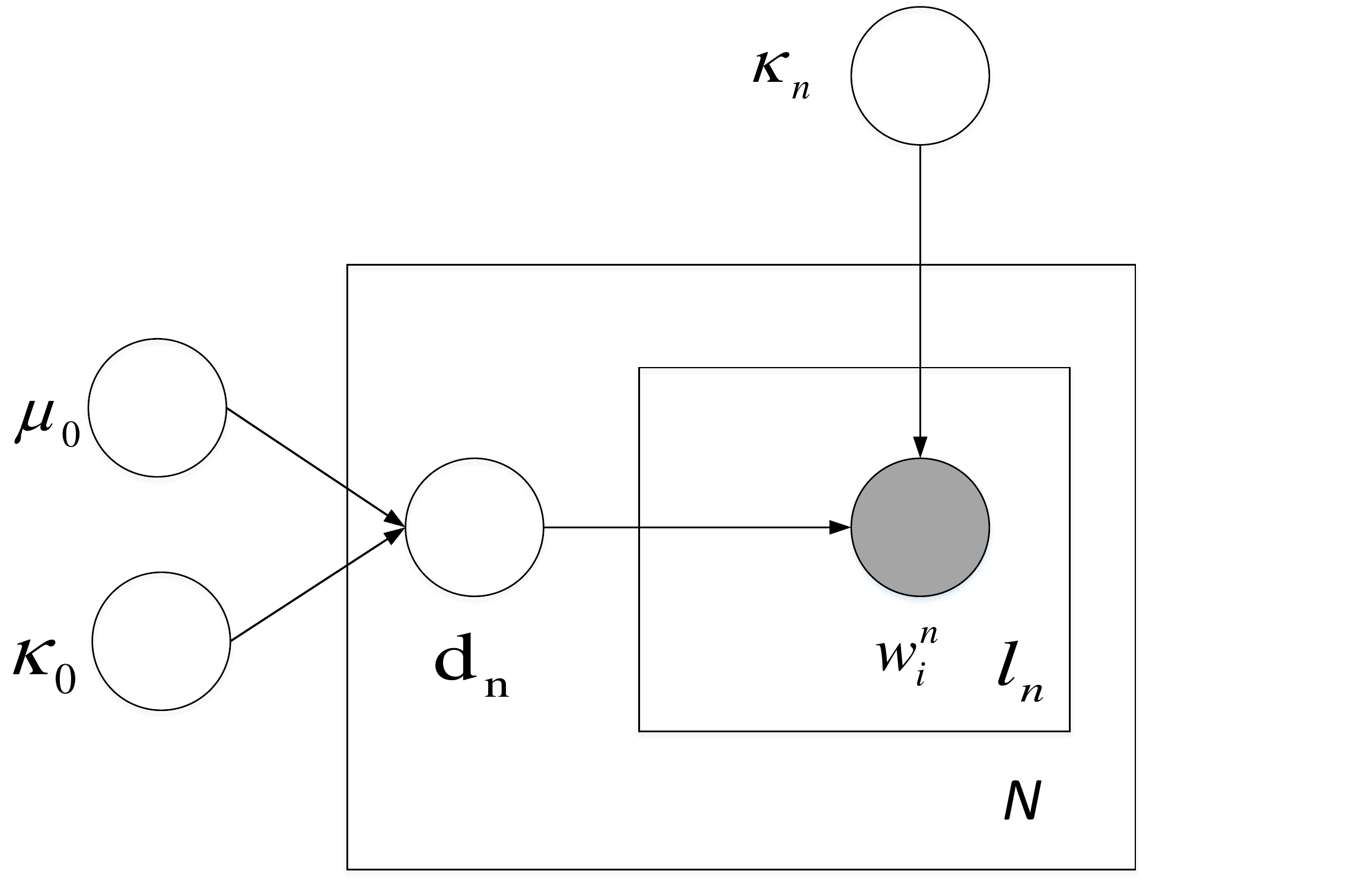}
		\caption{A graphical model representation of Spherical Paragraph Model (SPM). (The boxes are ``plates'' representing replicates; a shaded node is an observed variable; an unshaded node is a hidden variable.)}
		\label{fig:SPM}
\end{figure}

In this section, we describe our proposed SPM in detail, including the notations, the model definition, the inference and parameter estimation algorithms. Besides, we also provide some discussions on SPM as compared with existing advanced text representation methods.

\subsection{Notation}

Before presenting our model, we first introduce the notations used in this paper. Let $\boldsymbol{D}{=}\{d_1,\ldots,d_N\}$ denote a corpus of $N$ texts, where each text $d_n = (w_1^n,w_2^n,\ldots,w_{l_n}^n), n\in{1,2,\dots, N}$ is an $l_n$-length word sequence over the word vocabulary $\boldsymbol{V}$ of size $M$. Let $c_n$ denote all the words in text $d_n$. Each text $d \in \boldsymbol{D}$ and each word $ w \in \boldsymbol{V}$ is associated with a vector $\boldsymbol{d}\in \mathbb{R}^K$ and $\boldsymbol{w}\in \mathbb{R}^{K}$, respectively, where $K$ denotes the embedding dimensionality.

\subsection{Model Definition}

SPM is a probabilistic generative model over a text corpus based on BoWE. Specifically, each piece of text is first represented as a bag of $\ell_2$-normalized word vectors. Note that by normalization, the cosine similarity between word vectors is equal to the dot product between them, and all the word vectors lie on a unit hypersphere. SPM then assumes the following generative process of the corpus:
\begin{description}
\item For each text $d_n \in \boldsymbol{D}, n=1, 2, \dots,N$:
\item{(a)} Draw paragraph vector $\boldsymbol{d}_n \sim~$vMF$(\mu_0,\kappa_0)$
\item{(b)} For each word $w_i^n \in d_n, i=1,2,\dots,l_n$ :
\begin{description}
 \item Draw word vector $\boldsymbol{w}_i^n \sim~$vMF$(\boldsymbol{d}_n,\kappa_n)$
 \end{description}
\end{description}
where $\mu_0$ is the corpus mean direction, $\kappa_0$ controls the concentration of text vectors around $\mu_0$, and $\kappa_n$ controls the concentration of word vectors around the text vector $\boldsymbol{d}_n$. Figure~\ref{fig:SPM} provides the graphical model of the SPM.

As we can see from the above generative process, in SPM the text vectors in a corpus are determined by the corpus-wide prior distribution over the unit hypersphere, as well as the word vectors contained in the text. By using the vMF distribution, all the relations between these vectors are modeled by the dot product, which is equal to the cosine similarity measure between them (due to the $\ell_2$-normalization). As we known, cosine similarity is widely adopted in revealing semantic relatedness in previous neural word embedding methods \cite{mikolov2013efficient,mikolov2013distributed}.

Based on the above generative process, we can obtain the joint probability of the whole corpus as follows,
\begin{equation*}
   p(\boldsymbol{D}) {=} \prod_{n=1}^{N}  \int p(d_n|\mu_0,\kappa_0) \prod_{w_i^n\in d_n} p(w_{i}^n|{d_n},\kappa_n) d \boldsymbol{d}_n
\end{equation*}
where:
\begin{equation*}
P(w_i^n;d_n;\kappa_n)= e^{\kappa_n {\boldsymbol{d}_n}^{\mathrm{T}}\boldsymbol{w}_i^n} c_K{(\kappa_n)}
\end{equation*}

\subsection{Variational Inference}
\label{VI}
The key inferential problem that we need to solve in order to use SPM is that of computing the posterior distribution of the hidden text vector given its word vectors and the corpus prior:
\begin{equation*}
    p(d_n|c_n,\mu_0,\kappa_0,\kappa_n) {=} \frac{p(d_n,c_n|\mu_0,\kappa_0,\kappa_n)}{p(c_n|\mu_0,\kappa_0,\kappa_n)}
\end{equation*}

Unfortunately, this distribution is intractable to compute in general. Thus we develop an efficient variational inference algorithm to perform approximate inference in SPM.

The basic idea of convexity-based variational inference is to make use of Jensen's inequality \cite{jordan1999introduction} to obtain an adjustable lower bound on the log likelihood. We approximate the posterior by introducing an distinct vMF distribution for each document,
\vskip -0.1in
\begin{displaymath}
\begin{array}{rcl}
q(d_n) &\sim & $vMF$(.|\mu_n',\kappa_n')
\end{array}
\end{displaymath}
Here, $\mu_n',\kappa_n'$ are the free variational parameters. To approximate the posterior distribution of the latent variables, the mean-field approach finds the optimal parameters of the fully factorizable $q$ (\textit{i.e.}, $q(d_n)$) by maximizing the Evidence Lower Bound (ELBO),
\begin{equation*}
\begin{aligned}
\mathcal{L} &{=} E_q[\log P(\mathcal{D})] - \mathcal{H}(q)\\
&{=}E_q[\log P(\boldsymbol{D},\boldsymbol{V}|\mu_0,\kappa_0,\kappa_n)] - E_q[\log q(\boldsymbol{D})]\\
&{=}E_q[\log P(\boldsymbol{D}|\mu_0,\kappa_0)]+E_q[\log P(\boldsymbol{V}|\boldsymbol{D},\kappa_n)]\\
&{-}E_q[\log q(\boldsymbol{D})]
\end{aligned}
\end{equation*}

Note that the expectations in this expression are taken over the variational distribution $q$.
The posterior expectation of text vector $\boldsymbol{d}_n$ is given by,
\begin{equation*}
E_q[\boldsymbol{d}_n]=\mu_n'(\frac{I_{d/2}(\kappa_n')}{I_{d/2-1}(\kappa_n')})
\end{equation*}
where $\frac{I_{d/2}(\kappa_n')}{I_{d/2-1}(\kappa_n')}$ is a ratio of Bessel functions \cite{watson1995treatise} that differ in their order by just one.

Thus the optimizing values of the variational parameters $\mu_n'$ and $\kappa_n'$ are found by minimizing the KL divergence between the variational distribution $q$ and the true posterior $p(d_n|c_n,\mu_0,\kappa_0,\kappa_n)$. Optimizing the ELBO with respect to $\mu_n'$ and $\kappa_n'$, we have
\begin{equation*}
\kappa'_n = ||\kappa_0 \mu_0 + \sum_{i=1}^{l_n} \kappa_n \boldsymbol{w}_i^n||
\end{equation*}
\begin{equation*}
\mu'_n\!\!=\!\!\frac{\kappa_0 \mu_0 + \sum_{i=1}^{l_n} \kappa_n \boldsymbol{w}_i^n}{||\kappa_0 \mu_0 + \sum_{i=1}^{l_n} \kappa_n \boldsymbol{w}_i^n||}\! =\! \frac{\kappa_0 \mu_0 + \sum_{i=1}^{l_n} \kappa_n \boldsymbol{w}_i^n}{\kappa'_n}
\end{equation*}

\subsection{Parameter Estimation}
We use an empirical Bayes method for parameter estimation in our SPM model. As described above, variational inference provides us with a tractable lower bound on the log likelihood. We can thus find approximate empirical Bayes estimates via an alternating variational EM procedure that maximizes the lower bound with respect to the variational parameters $\mu_n'$ and $\kappa_n'$. Then, for fixed values of the variational parameters, we maximize the lower bound with respect to the model parameters $\mu_0,\kappa_0$ and $\kappa_n$. The variational EM algorithm is as follows:
\begin{itemize}[leftmargin=*]
\item (E-step) For each text, find the optimizing values of the variational parameters {$\mu_n',\kappa_n'$}, as described in the previous section \ref{VI}.
\item (M-step) Maximize the lower bound with respect to the model parameters $\mu_0,\kappa_0$ and $\kappa_n$.
\end{itemize}
These two steps are repeated until the lower bound on the log likelihood converges.
The M-step update for $\mu_0, \kappa_0$ are given by,
\begin{equation*}
\mu_0 = \frac{\sum_{n=1}^{N} E_q[\boldsymbol{d}_n]}{||\sum_{n=1}^{N} E_q[\boldsymbol{d}_n]||}
\end{equation*}
\begin{equation*}
\kappa_0 = \frac{{\bar r}K-{\bar r}^3}{1-{\bar r}^2}~~
\text{where}~~
\bar{r} = \frac{||\sum_{n=1}^{N}E_q[\boldsymbol{d}_n]||}{N}
\end{equation*}
The M-step update for $\kappa_n$ is given by,
\begin{equation*}
\kappa_n = \frac{{\bar r}K-{\bar r}^3}{1-{\bar r}^2}~~
\text{where}~~
\bar{r} = \frac{  E_q[\boldsymbol{d}_n] \sum_{i=1}^{l_n} {\boldsymbol{w}_i^n}^{\mathrm{T}}}{l_n}
\end{equation*}

\subsection{Model Discussion}
SPM is a probabilistic generative model based on BoWE for text representation. As it bridges two well-known branches in text representation methods, namely the probabilistic generative models and neural embedding models, here we compare SPM with these two types of methods to show its benefits.

Probabilistic generative models, also called probabilistic topic models (\textit{e.g.}, PLSI and LDA), are advanced text modeling approaches. By assuming a generative process of the texts under a probabilistic framework, these methods usually have sound theoretical foundation and good model interpretability. However, there are two major problems in traditional topic models: (1)
As built upon the BoW representation, traditional topic methods do not leverage the rich semantic relatedness of the words, and learn the text representations purely based on the word-by-text co-occurrence information; (2) There is an explicit topic layer in these models to guide the word clustering. The topic number is usually heuristically defined \textit{a prior} which may lead to non-optimal word clustering.
As we can see, SPM enjoys the merits of good interpretability as a probabilistic generative model. Meanwhile, SPM can avoid the arbitrary definition of topic numbers by eliminating the topic layer, while allows unlimited hidden topics (\textit{i.e.}, word clusters) learned by any prior neural word embedding models based on very large corpus.

As compared with neural embedding models, here we take the state-of-the-art PV model as an example. The PV model can also be viewed as a probabilistic model based on its prediction definition. However, from the probabilistic view, PV is not a full Bayesian model and suffers a similar problem as PLSI that it provides no model on text vectors.
Therefore, texts from the same corpus are assumed to be independent from each other and no corpus-wide constraint is employed in text modeling. Moreover, it is unclear how to infer the representations for texts outside of the training set with the learned model. Although PV makes itself as an optimization problem so that one can learn representations for new texts anyway, it loses the sound probabilistic foundation in that way. In contrary, SPM solves this problem by defining a complete Bayesian model. In this way, it can not only leverage corpus-wide information to help constrain the text vectors, but also infer the representations of unseen texts based on the learned model, at the expense of the usage of an approximate variational method.

\section{Experiments}
\label{experiments}
In this section, we conduct experiments to verify the effectiveness of SPM based on two text classification tasks.

\subsection{Baselines}

\begin{itemize}[leftmargin=*]
\item \textbf{Bag-of-Words}. The Bag-of-Words model (BoW) \cite{harris1954distributional} represents each text as a bag of words using \textit{tf} as the weighting scheme. We select top $5,000$ words according to \textit{tf} scores as discriminative features.
\item \textbf{LSI} and \textbf{LDA}. LSI \cite{deerwester1990indexing} maps both texts and words to lower-dimensional representations
using SVD decomposition. In LDA \cite{blei2003latent}, each word within a text is modeled as a finite mixture over an set of topics. We use the vanilla LSI and LDA in the gensim library\footnote{\url{http://radimrehurek.com/gensim/}} with topic number set as 50.
\item \textbf{movMF} and \textbf{SAM}. The movMF\footnote{\url{https://github.com/mrouvier/movMF}} \cite{banerjee2005clustering} is the mixture of von-Mises Fisher clustering with soft assignments. The SAM\footnote{\url{https://github.com/austinwaters/py-sam}} \cite{reisinger2010spherical} is a class of topic models that represent data using directional distributions on the unit hypersphere. The topic numbers are both 50.
\item \textbf{cBow}. We use average pooling to compose a text vector from a set of word vectors \cite{mikolov2013efficient}, where the dimension of text vectors is set as 50.
\item \textbf{PV}. Paragraph Vector \cite{le2014distributed} is an unsupervised model to learn distributed representations of words and texts. We implement PV-DBOW and PV-DM model initialized with 50-dimension word embeddings due to the original code has not been released.
\item \textbf{skip-thought} and \textbf{FastSent}. skip-thought\footnote{\url{https://github.com/ryankiros/skip-thoughts}} \cite{kiros2015skip} encodes a sentence to predict sentences around it using 2400-dimension vector representation. FastSent\footnote{\url{https://github.com/fh295/SentenceRepresentation}} \cite{hill2016learning} is a simple additive sentence model designed to exploit the same signal, but at much lower computational expense under 100 dimension.
\end{itemize}
\subsection{Setup}
We perform experiments on two text classification tasks: topical classification and sentiment analysis. We utilize 50-dimension word embeddings trained on Wikipedia with word2vec\footnote{\url{https://code.google.com/p/word2vec/}}. The corpus in total has $3,035,070$ articles and about 1 billion tokens. The vocabulary size is about $400,000$. The vectors are post-processed to have unit $\ell_2$-norm. In our model, text vectors are randomly initialized with values uniformly distributed in the range of [-0.5, +0.5] with 50-dimension and then $\ell_2$-normalized, $\kappa_0$ is intialized as 1500 and $\kappa_n$ are randomly initialized with values uniformly distributed in the range of [1000, 1500]. Through our experiments, we use support vector machines (SVM)\footnote{\url{http://www.csie.ntu.edu.tw/~cjlin/libsvm/}} as the classifier. 
Preprocessing steps were applied to all datasets: words were lowercased, non-English characters and stop words occurrence in the training set are removed. If explicit split of train/test is not provided, we use 10-fold cross-validation instead.

\subsection{Topical Classification}

We used two standard topical classification corpora: the 20Newsgroups\footnote{\url{http://qwone.com/~jason/20Newsgroups/}} and the Reuters corpus \footnote{\url{http://www.nltk.org/book/ch02.html}}. The 20Newsgroups contains about $20,000$ newsgroup documents harvested from 20 different Usenet newsgroups, with about $1,000$ documents from each newsgroup. Following \citeauthor{banerjee2007topic} \shortcite{banerjee2007topic}, three subsets of 20News are used for evaluation: (1) \textbf{news-20-different} consists of three newsgroups that cover different topics (\textit{rec.sport.baseball}, \textit{sci.space} and \textit{alt.atheism}); (2) \textbf{news-20-similar} consists of three newsgroups on the more similar topics (\textit{rec.sport.baseball}, \textit{talk.politics.guns} and \textit{talk.politics.misc}); (3) \textbf{news-20-same} consists of three newsgroups on the highly related topics (\textit{comp.os.ms-windows.misc}, \textit{comp.windows.x} and \textit{comp.graphics}). The Reuters contains $10,788$ documents, where each document is assigned to one or more categories. Documents appearing in two or more categories were removed and we selected the largest 10 categories, leaving $8,025$ documents in total.
\begin{table}[t]
  \renewcommand{\arraystretch}{0.98}
  \setlength\tabcolsep{4pt}
  \caption{Classification accuracies (\%) of different models on topical classification.}
  \centering
  \begin{tabular}{l c c c c c c} \toprule
   Model &  different & similar & same & Reuters\\ \midrule
    BoW   & 91.4 & 81.8 & 75.6 & \textbf{95.4}\\
    LSI & 85.2& 80.1 & 68.2 & 93.1\\
    LDA & 73.3 & 67.5 & 56.7 & 89.6 \\
    movMF  & 71.4 & 64.5 & 59.4 & 87.1 \\
    SAM & 88.6 & 81.2 & 70.5 & 88.2  \\
    cBow & 91.6 & 81.6 &75.9 & 91.8\\
    PV-DBOW & 91.4& 80.2 &  \textbf{76.2} & 89.6 \\
    PV-DM & 91.5 & 80.8 &  76.1 & 90.4 \\
    FastSent & 89.6 & 80.1 & 61.5 & 89.4 & \\
    uni-skip & 86.4 & 77.8 & 59.2 & 77.4 & \\
    SPM & \textbf{91.8} & \textbf{82.0}  & 70.0 & 93.2 \\\bottomrule
     \end{tabular}
     \label{table:20news}
\end{table}

\textbf{Results} Table \ref{table:20news} shows the evaluation results on topical classification. We have the following observations: (1) The BoW representation, although simple,  can achieve surprising accuracy using much larger dimensionality (\textit{i.e.}, $5,000$ dimension). Meanwhile, our SPM, using only 50-dimension text vector, can achieve slightly better or comparable performance as BoW. (2) As compared with the text representation methods built upon BoW (\textit{i.e.}, LSI, LDA, movMF and SAM), SPM can outperform these methods almost. The results indicate that learning text representations over BoWE can in general achieve better performances than that over BoW by involving rich semantics between words. (3) Comparing with the three BoWE based representation methods, namely cBow, PV-DBOW and PV-DM, we find our SPM can outperform them on three out of four datasets. Recall that in cBow, PV-DBOW and PV-DM, texts in a corpus are actually assumed to be independent from each other.  These results indicate that by modeling texts under a sound probabilistic generative framework, SPM can well leverage the corpus-wide information to help improve the text representation. (4) Compared with FastSent and uni-skip, SPM can outperform both of them over the four datasets. It seems that FastSent and uni-skip, which were proposed for short texts (\textit{i.e.}, sentences) modeling originally, cannot work well on long texts. 
\begin{figure*}[t]
  \begin{center}
    \ref{named}
    \begin{tikzpicture}[scale=0.75]
      \begin{axis}[
        enlarge x limits=0.2,
        xlabel=Subj,
        ylabel=test accuracy $\%$,
        ymin=60, ymax=100,
        xticklabels={50,100,300},
        xtick={1,2,3},
        legend style={
          font=\small,
          draw=none,
          legend columns=-1,
          at={(0.5,1.1)},
          anchor=north,
          /tikz/every even column/.append style={column sep=12mm}
        },
        ybar=1pt,
        bar width=8pt,
        ymajorgrids,
        major grid style={draw=gray3},
        y axis line style={opacity=0},
        tickwidth=0pt,
        legend entries = {LSI,LDA,cBow,PV-DBOW,PV-DM,SPM},
        legend to name=named]
        ]
        \addplot [draw=none, fill=green3]
        coordinates {
          (1, 85.4)
          (2, 87.6)
          (3, 89.4)};
        \addplot [draw=none, fill=green2]
        coordinates {
          (1, 72.7)
          (2, 71)
          (3, 79.8)};
        \addplot [draw=none, fill=purple1]
        coordinates {
          (1, 90.8)
          (2, 91.2)
          (3, 92.3)};
        \addplot [draw=none, fill=yellow1]
        coordinates {
          (1, 90.1)
          (2, 90.8)
          (3, 91.4)};
        \addplot [draw=none, fill=pink1]
        coordinates {
          (1, 90.2)
          (2, 91.1)
          (3, 92.0)};
        \addplot [draw=none, fill=orange1]
        coordinates {
          (1, 92.5)
          (2, 92.6)
          (3, 92.8)};
      \end{axis}
    \end{tikzpicture}
    \begin{tikzpicture}[scale=0.75]
      \begin{axis}[
        enlarge x limits=0.2,
        xlabel=MR,
        ylabel=test accuracy $\%$,
        ymin=50, ymax=80,
        xticklabels={50,100,300},
        xtick={1,2,3},
        ybar=1pt,
        bar width=8pt,
        ymajorgrids,
        major grid style={draw=gray3},
        y axis line style={opacity=0},
        tickwidth=0pt,
        ]
        \addplot [draw=none, fill=green3]
        coordinates {
          (1, 64.2)
          (2, 66.8)
          (3, 69.4)};
        \addplot [draw=none, fill=green2]
        coordinates {
          (1, 58.2)
          (2, 61.6)
          (3, 65.2)};
        \addplot [draw=none, fill=purple1]
        coordinates {
          (1, 74.4)
          (2, 75.1)
          (3, 77.8)};
        \addplot [draw=none, fill=yellow1]
        coordinates {
          (1, 73.9)
          (2, 75.9)
          (3, 77.6)};
        \addplot [draw=none, fill=pink1]
        coordinates {
          (1, 74.4)
          (2, 75.8)
          (3, 78.5)};
        \addplot [draw=none, fill=orange1]
        coordinates {
          (1, 75.0)
          (2, 76.0)
          (3, 79.1)};

      \end{axis}
    \end{tikzpicture}

  \end{center}
  \caption{Classification accuracies on sentiment analysis tasks under different dimensionality.}
  \label{fig:Dimen}
\end{figure*}
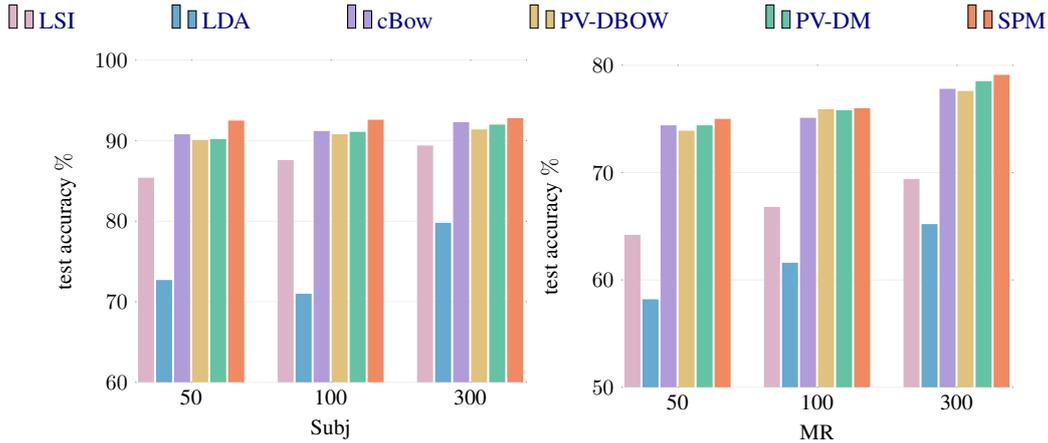

\subsection{Sentiment Analysis}
We run the sentiment classification experiments on two publicly available datasets.
\begin{itemize}[leftmargin=*]
\item \textbf{Subj}, Subjectivity dataset \cite{pang2004sentimental}\footnote{\url{http://www.cs.cornell.edu/people/pabo/movie-review-data/}} which contains $5,000$ subjective instances and $5,000$ objective instances. The task is to classify a sentence as being subjective or objective;
\item \textbf{MR}, Movie reviews \cite{pang2005seeing} with one sentence per review. There are $5,331$ positive sentences and $5,331$ negative sentences. Classification involves detecting positive/negative reviews.
\end{itemize}

\textbf{Results} Table \ref{table:senti} shows the evaluation results on two datasets. We have the following observations: (1) SPM can outperform all the baseline methods on the Subj dataset. This indicates that SPM can capture better semantic representations of texts using a probabilistic generative model over BoWE. (2) SPM can also outperform all the baseline methods except uni-skip on the MR dataset. Note that skip-thought uses 2400-dimension sentence representation while SPM only uses 50-dimension vector. However, SPM can still achieve similar performance as uni-skip on the MR dataset even with much less model parameters. 
\vskip -0.15in
\begin{table}[h]
  \renewcommand{\arraystretch}{1}
  \setlength\tabcolsep{10pt}
  \centering
  \caption{Classification accuracies (\%) of different models on sentiment analysis.}
  \begin{tabular}{l c c c } \toprule
   Model   & Subj & MR & \\ \midrule
    BoW & 89.5 & 74.3 &  \\
    LSI   & 85.4 & 64.2 &  \\
    LDA  & 72.7 & 58.2 & \\
    movMF & 67.6 & 53.4 & \\
    SAM & 74.2 & 61.8 & \\
    cBow  & 90.8 & 74.4 & \\
    PV-DBOW  & 90.1 & 73.9 &\\
    PV-DM  & 90.4 & 74.4 &  \\
    FastSent  & 88.7 & 70.8 &   \\
    uni-skip & 92.1 & \textbf{75.5}& \\\midrule
    SPM & \textbf{92.5} & 75.0 &  \\ \bottomrule
  \end{tabular}
  \label{table:senti}
\end{table}

We conduct evaluations over different dimensions (\textit{i.e.}, 50, 100, 300) to see the impact of the dimensionality on different models. For cBow, PV and SPM, we utilize 50, 100 and 300 dimensional word embeddings trained on Wikipedia using word2vec. For LSI and LDA, we set the topic numbers as 50, 100 and 300 for comparison. Figure \ref{fig:Dimen} shows the results on the two different datasets. As we can see, with the increase of the dimensionality, all the models can improve their performance while SPM can consistently outperform all the other baselines. Moreover, we can find that the SPM model under dimensionality 100 can already beat the uni-skip under dimensionality 2400 (76.0\% vs 75.5\%) on the MR dataset.

\section{Conclusion}
\label{conclusion}

In this paper, we propose the SPM, a novel generative model based on BoWE for text modeling. The SPM is a full Bayesian framework which models the generation of both the text vectors and word vectors, where the vMF distribution is employed to capture the directional relations between these vectors. SPM has good probabilistic interpretability and can fully leverage the rich semantics of words, the word co-occurrence information as well as the corpus-wide information to help the representation learning. The experimental results demonstrate that SPM can achieve new state-of-the-art performances on several topical classification and sentiment analysis tasks.

For the future work, we would like to explore the possibility to jointly learn word and text vectors in SPM. One idea is to leverage the word vectors learned from other large corpus as the initialization, and fine-tune them on the training data under SPM. Moreover, word order information is often critical in capturing the meaning of texts. We would also try to accommodate n-grams in the generative process to enhance the model ability. We may also test SPM on other text processing tasks to verify its generalization ability.

\bibliographystyle{emnlp_natbib}
\bibliography{emnlp2017}

\begin{thebibliography}{30}
\expandafter\ifx\csname natexlab\endcsname\relax\def\natexlab#1{#1}\fi

\bibitem[{Banerjee and Basu(2007)}]{banerjee2007topic}
Arindam Banerjee and Sugato Basu. 2007.
\newblock Topic models over text streams: A study of batch and online
  unsupervised learning.
\newblock In \emph{SDM}, volume~7, pages 437--442. SIAM.

\bibitem[{Banerjee et~al.(2005)Banerjee, Dhillon, Ghosh, and
  Sra}]{banerjee2005clustering}
Arindam Banerjee, Inderjit~S Dhillon, Joydeep Ghosh, and Suvrit Sra. 2005.
\newblock Clustering on the unit hypersphere using von mises-fisher
  distributions.
\newblock \emph{Journal of Machine Learning Research}, 6(Sep):1345--1382.

\bibitem[{Blei et~al.(2003)Blei, Ng, and Jordan}]{blei2003latent}
David~M Blei, Andrew~Y Ng, and Michael~I Jordan. 2003.
\newblock Latent dirichlet allocation.
\newblock \emph{Journal of machine Learning research}, 3(Jan):993--1022.

\bibitem[{Clinchant and Perronnin(2013)}]{clinchant2013aggregating}
St{\'e}phane Clinchant and Florent Perronnin. 2013.
\newblock Aggregating continuous word embeddings for information retrieval.
\newblock In \emph{Proceedings of the Workshop on Continuous Vector Space
  Models and their Compositionality}, pages 100--109.

\bibitem[{Deerwester et~al.(1990)Deerwester, Dumais, Furnas, Landauer, and
  Harshman}]{deerwester1990indexing}
Scott Deerwester, Susan~T Dumais, George~W Furnas, Thomas~K Landauer, and
  Richard Harshman. 1990.
\newblock Indexing by latent semantic analysis.
\newblock \emph{Journal of the American society for information science},
  41(6):391.

\bibitem[{Fisher(1953)}]{fisher1953dispersion}
Ronald Fisher. 1953.
\newblock Dispersion on a sphere.
\newblock In \emph{Proceedings of the Royal Society of London A: Mathematical,
  Physical and Engineering Sciences}, volume 217, pages 295--305. The Royal
  Society.

\bibitem[{Harris(1954)}]{harris1954distributional}
Zellig~S Harris. 1954.
\newblock Distributional structure.
\newblock \emph{Word}, 10(2-3):146--162.

\bibitem[{Hill et~al.(2016)Hill, Cho, and Korhonen}]{hill2016learning}
Felix Hill, Kyunghyun Cho, and Anna Korhonen. 2016.
\newblock Learning distributed representations of sentences from unlabelled
  data.
\newblock In \emph{NAACL-HLT}.

\bibitem[{Hofmann(1999)}]{hofmann1999probabilistic}
Thomas Hofmann. 1999.
\newblock Probabilistic latent semantic indexing.
\newblock In \emph{Proceedings of the 22nd annual international ACM SIGIR
  conference on Research and development in information retrieval}, pages
  50--57. ACM.

\bibitem[{Jaakkola et~al.(1999)Jaakkola, Haussler
  et~al.}]{jaakkola1999exploiting}
Tommi~S Jaakkola, David Haussler, et~al. 1999.
\newblock Exploiting generative models in discriminative classifiers.
\newblock \emph{Advances in neural information processing systems}, pages
  487--493.

\bibitem[{Jordan et~al.(1999)Jordan, Ghahramani, Jaakkola, and
  Saul}]{jordan1999introduction}
Michael~I Jordan, Zoubin Ghahramani, Tommi~S Jaakkola, and Lawrence~K Saul.
  1999.
\newblock An introduction to variational methods for graphical models.
\newblock \emph{Machine learning}, 37(2):183--233.

\bibitem[{Jupp and Mardia(1989)}]{jupp1989unified}
PE~Jupp and KV~Mardia. 1989.
\newblock A unified view of the theory of directional statistics, 1975-1988.
\newblock \emph{International Statistical Review/Revue Internationale de
  Statistique}, pages 261--294.

\bibitem[{Kim(2014)}]{kim2014convolutional}
Yoon Kim. 2014.
\newblock Convolutional neural networks for sentence classification.
\newblock In \emph{EMNLP}, pages 1746--1751.

\bibitem[{Kiros et~al.(2015)Kiros, Zhu, Salakhutdinov, Zemel, Urtasun,
  Torralba, and Fidler}]{kiros2015skip}
Ryan Kiros, Yukun Zhu, Ruslan~R Salakhutdinov, Richard Zemel, Raquel Urtasun,
  Antonio Torralba, and Sanja Fidler. 2015.
\newblock Skip-thought vectors.
\newblock In \emph{Advances in neural information processing systems}, pages
  3294--3302.

\bibitem[{Le and Mikolov(2014)}]{le2014distributed}
Quoc~V Le and Tomas Mikolov. 2014.
\newblock Distributed representations of sentences and documents.
\newblock In \emph{ICML}, volume~14, pages 1188--1196.

\bibitem[{Lee and Seung(1999)}]{lee1999learning}
Daniel~D Lee and H~Sebastian Seung. 1999.
\newblock Learning the parts of objects by non-negative matrix factorization.
\newblock \emph{Nature}, 401(6755):788--791.

\bibitem[{Manning et~al.(1999)Manning, Sch{\"u}tze
  et~al.}]{manning1999foundations}
Christopher~D Manning, Hinrich Sch{\"u}tze, et~al. 1999.
\newblock \emph{Foundations of statistical natural language processing}, volume
  999.
\newblock MIT Press.

\bibitem[{Mardia and Jupp(2009)}]{mardia2009directional}
Kanti~V Mardia and Peter~E Jupp. 2009.
\newblock \emph{Directional statistics}, volume 494.
\newblock John Wiley \& Sons.

\bibitem[{Mikolov et~al.(2013{\natexlab{a}})Mikolov, Chen, Corrado, and
  Dean}]{mikolov2013efficient}
Tomas Mikolov, Kai Chen, Greg Corrado, and Jeffrey Dean. 2013{\natexlab{a}}.
\newblock Efficient estimation of word representations in vector space.
\newblock \emph{arXiv preprint arXiv:1301.3781}.

\bibitem[{Mikolov et~al.(2013{\natexlab{b}})Mikolov, Sutskever, Chen, Corrado,
  and Dean}]{mikolov2013distributed}
Tomas Mikolov, Ilya Sutskever, Kai Chen, Greg~S Corrado, and Jeff Dean.
  2013{\natexlab{b}}.
\newblock Distributed representations of words and phrases and their
  compositionality.
\newblock In \emph{Advances in neural information processing systems}, pages
  3111--3119.

\bibitem[{Pang and Lee(2004)}]{pang2004sentimental}
Bo~Pang and Lillian Lee. 2004.
\newblock A sentimental education: Sentiment analysis using subjectivity
  summarization based on minimum cuts.
\newblock In \emph{Proceedings of the 42nd annual meeting on Association for
  Computational Linguistics}, page 271. Association for Computational
  Linguistics.

\bibitem[{Pang and Lee(2005)}]{pang2005seeing}
Bo~Pang and Lillian Lee. 2005.
\newblock Seeing stars: Exploiting class relationships for sentiment
  categorization with respect to rating scales.
\newblock In \emph{Proceedings of the 43rd annual meeting on association for
  computational linguistics}, pages 115--124. Association for Computational
  Linguistics.

\bibitem[{Pennington et~al.(2014)Pennington, Socher, and
  Manning}]{pennington2014glove}
Jeffrey Pennington, Richard Socher, and Christopher~D Manning. 2014.
\newblock Glove: Global vectors for word representation.
\newblock In \emph{EMNLP}, volume~14, pages 1532--1543.

\bibitem[{Reisinger et~al.(2010)Reisinger, Waters, Silverthorn, and
  Mooney}]{reisinger2010spherical}
Joseph Reisinger, Austin Waters, Bryan Silverthorn, and Raymond~J Mooney. 2010.
\newblock Spherical topic models.
\newblock In \emph{Proceedings of the 27th international conference on machine
  learning (ICML-10)}, pages 903--910.

\bibitem[{Salton and McGill(1986)}]{salton1986introduction}
Gerard Salton and Michael~J McGill. 1986.
\newblock Introduction to modern information retrieval.

\bibitem[{Socher et~al.(2013)Socher, Perelygin, Wu, Chuang, Manning, Ng, and
  Potts}]{socher2013recursive}
Richard Socher, Alex Perelygin, Jean~Y Wu, Jason Chuang, Christopher~D Manning,
  Andrew~Y Ng, and Christopher Potts. 2013.
\newblock Recursive deep models for semantic compositionality over a sentiment
  treebank.
\newblock In \emph{Proceedings of the conference on empirical methods in
  natural language processing (EMNLP)}, volume 1631, page 1642. Citeseer.

\bibitem[{Sutskever et~al.(2011)Sutskever, Martens, and
  Hinton}]{sutskever2011generating}
Ilya Sutskever, James Martens, and Geoffrey~E Hinton. 2011.
\newblock Generating text with recurrent neural networks.
\newblock In \emph{Proceedings of the 28th International Conference on Machine
  Learning (ICML-11)}, pages 1017--1024.

\bibitem[{Vulic and Moens(2013)}]{vulic2013cross}
Ivan Vulic and Marie-Francine Moens. 2013.
\newblock Cross-lingual semantic similarity of words as the similarity of their
  semantic word responses.
\newblock In \emph{Proceedings of the 2013 Conference of the North American
  Chapter of the Association for Computational Linguistics: Human Language
  Technologies (NAACL-HLT 2013)}, pages 106--116. ACL.

\bibitem[{Watson(1995)}]{watson1995treatise}
George~Neville Watson. 1995.
\newblock \emph{A treatise on the theory of Bessel functions}.
\newblock Cambridge university press.

\bibitem[{Zhong and Ghosh(2005)}]{zhong2005generative}
Shi Zhong and Joydeep Ghosh. 2005.
\newblock Generative model-based document clustering: a comparative study.
\newblock \emph{Knowledge and Information Systems}, 8(3):374--384.

\end{thebibliography}

\end{document}